\documentclass{article}

\usepackage{PRIMEarxiv}

\usepackage[utf8]{inputenc} % allow utf-8 input
\usepackage[T1]{fontenc}    % use 8-bit T1 fonts
\usepackage{hyperref}       % hyperlinks
\usepackage{url}            % simple URL typesetting
\usepackage{booktabs}       % professional-quality tables
\usepackage{amsfonts}       % blackboard math symbols
\usepackage{nicefrac}       % compact symbols for 1/2, etc.
\usepackage{microtype}      % microtypography
\usepackage{lipsum}
\usepackage{fancyhdr}       % header
\usepackage{graphicx}       % graphics
\graphicspath{{media/}}     % organize your images and other figures under media/ folder
\usepackage{amsfonts}
\usepackage{siunitx}
\usepackage{caption}
\usepackage{subcaption}
\usepackage{graphicx}
\usepackage{multirow}
\usepackage{hyperref}
\usepackage{amsmath}

\title{Reinforcement Learning for Safe Autonomous Two Device Navigation of Cerebral Vessels in Mechanical Thrombectomy
\thanks{\textit{\underline{Citation}}: 
\textbf{Robertshaw, H., Jackson, B., Wang., J., et al. Reinforcement learning for safe autonomous two device navigation of cerebral vessels in mechanical thrombectomy. Int J CARS (2025). https://doi.org/10.1007/s11548-025-03339-8}} 
}

\author{
  Harry Robertshaw, Benjamin Jackson, Jiaheng Wang, Hadi Sadati, Alejandro Granados, Thomas C Booth \\
  School of Biomedical Engineering \& Imaging Sciences \\
  Kings College London \\
  London\\
   \And
  Lennart Karstensen \\
  AIBE \\
  Friedrich-Alexander University Erlangen-Nürnberg \\
  Erlangen\\
}

\begin{document}
\maketitle

\begin{abstract}
    \textbf{Purpose:} Autonomous systems in mechanical thrombectomy (MT) hold promise for reducing procedure times, minimizing radiation exposure, and enhancing patient safety. However, current reinforcement learning (RL) methods only reach the carotid arteries, are not generalizable to other patient vasculatures, and do not consider safety. We propose a safe dual-device RL algorithm that can navigate beyond the carotid arteries to cerebral vessels.
    
    \textbf{Methods:} We used the Simulation Open Framework Architecture to represent the intricacies of cerebral vessels, and a modified Soft Actor-Critic RL algorithm to learn, for the first time, the navigation of micro-catheters and micro-guidewires. We incorporate patient safety metrics into our reward function by integrating guidewire tip forces. Inverse RL is used with demonstrator data on 12 patient-specific vascular cases.

    \textbf{Results:} Our simulation demonstrates successful autonomous navigation within unseen cerebral vessels, achieving a 96\% success rate, 7.0\,s procedure time, and 0.24~\si{\newton} mean forces, well below the proposed 1.5~\si{\newton} vessel rupture threshold. 
    
    \textbf{Conclusion:} To the best of our knowledge, our proposed autonomous system for MT two-device navigation reaches cerebral vessels, considers safety, and is generalizable to unseen patient-specific cases for the first time. We envisage future work will extend the validation to vasculatures of different complexity and on \textit{in vitro} models. While our contributions pave the way towards deploying agents in clinical settings, safety and trustworthiness will be crucial elements to consider when proposing new methodology.
\end{abstract}

% keywords can be removed

\section{Introduction}

    Ischemic stroke causes 3.48~million global deaths annually, and accounts for \$36.5 billion in direct medical costs in the U.S. alone \cite{Vos2020, Tsao2023}. Mechanical thrombectomy (MT) has emerged as a standard treatment for acute ischemic stroke resulting from large vessel occlusion, providing improved functional outcomes and lower mortality rates compared to medical treatment alone \cite{Bendszus2023, Nogueira2018}. In MT, a guidewire is used to navigate a guide catheter from the femoral or radial artery to the internal carotid artery (ICA). An `access catheter' is then advanced ahead of the guide catheter for vessel branch access. Once the access catheter is within the ICA, the guide catheter is advanced to make a stable platform. The guidewire and access catheter are then retracted, and a micro-guidewire within a micro-catheter is passed through the stable guide catheter and navigated to the target thrombus site which is typically within the M1 segment of the middle cerebral artery (MCA). The final step is to remove the micro-guidewire and exchange it for a stent-retriever to remove the thrombus thereby restoring blood flow to the brain.
    
    MT is most effective when performed early, ideally within 24 hours of stroke onset \cite{Nogueira2018}. However, only 3.1\% of stroke patients in the UK receive MT, despite 10\% being eligible \cite{McMeekin2017, SSNAP2023}. Limited access to MT centers, long travel times, and procedural risks like vessel damage and radiation exposure pose significant challenges for the delivery of MT \cite{Hausegger2001, Klein2009}. Robotic surgical systems offer a potential solution by increasing access to MT and reducing operator risks. Tele-operated MT could allow centralized, highly-trained specialists to perform the procedure remotely, improving access in underserved regions. Alternatively, assistive robots in peripheral hospitals could help less experienced operators perform MT safely. In either scenario, integrating AI into robotic systems may further enhance procedural efficiency and safety, especially for less experienced users (e.g., interventional radiologists who are not used to performing neurointerventional procedures) \cite{Crinnion2022, Jackson2023}.

    A recent review on the autonomous navigation of endovascular interventions found that 71\% (10/14) of studies performed experiments in the blood vessels within or around the heart \cite{Robertshaw2023}. More recently, a combined reward model using demonstrator data in inverse reinforcement learning (IRL) has been shown to provide improved success rates over a traditional dense reward function in the autonomous navigation of devices from the common iliac artery to the ICA, however, this work was limited to training and testing on exactly the same vasculature \cite{Robertshaw2024}. While reinforcement learning (RL) models have been tested on unseen vessel structures previously (for aortic arch navigation), these structures are typically simplified and computer generated \cite{Karstensen2023, Scarponi2024}. Furthermore, while 29\% (4/14) of studies reviewed in \cite{Robertshaw2023} measure the forces during navigation, there is no evidence of excessive force penalization. This is important to consider because (1) excessive vessel wall contact forces can induce vasoconstriction and damage, leading to reactive intimal proliferation chronically, or more importantly, dissection-induced distal embolization and infarction in the acute setting \cite{Takashima2007}; (2) it is known that robotic endovascular manipulators can reduce vessel wall contact forces \cite{RafiiTari2016}; and (3) patient and public involvement exercises on robotic MT showed concern about micro-guidewire safety - AI assistance was a potential solution for safety. 

    The aim of this study was to train an RL model capable of autonomously navigating a micro-guidewire and micro-catheter from the ICA to the MCA in unseen patient vasculatures while incorporating device tip forces. The primary objective was to demonstrate in an MT environment that RL can successfully perform this navigation in complex, unseen patient vasculatures. The secondary objectives were: 1) to assess whether adding force feedback in the reward function improves surrogate measures of patient safety and/or affects navigation performance, and 2) to validate previous findings that a combined reward model enhances navigation performance. This study used two-dimensional Cartesian tracking co-ordinates as inputs to the RL agent. Using this proposed method, it is plausible that effective clinical translation would be enacted using image-tracking to obtain the guidewire and catheter tip points from live fluoroscopic images during MT \cite{Eyberg2022}. This would then provide the inputs (i.e., two-dimensional co-ordinates) needed for the RL agent to navigate the vasculature. Furthermore, as tip forces are only required during training as part of the reward function, there would be no need to measure forces during the intervention. As the trained RL agent doesn't need a reward input, no additional equipment or software would be required in the clinic.
    
    Our contributions are as follows: 1) we applied RL to train autonomous navigation models for micro-guidewire and micro-catheter in cerebral vessels, specifically targeting MT, for the first time, 2) we tested the model on realistic, unseen patient vasculatures, and 3) we introduced penalization for instrument forces in the RL model to enhance patient safety.

\section{Methods}

    \subsection{Navigation task}

\begin{figure}[]
            \centering
            \includegraphics[width=9cm]{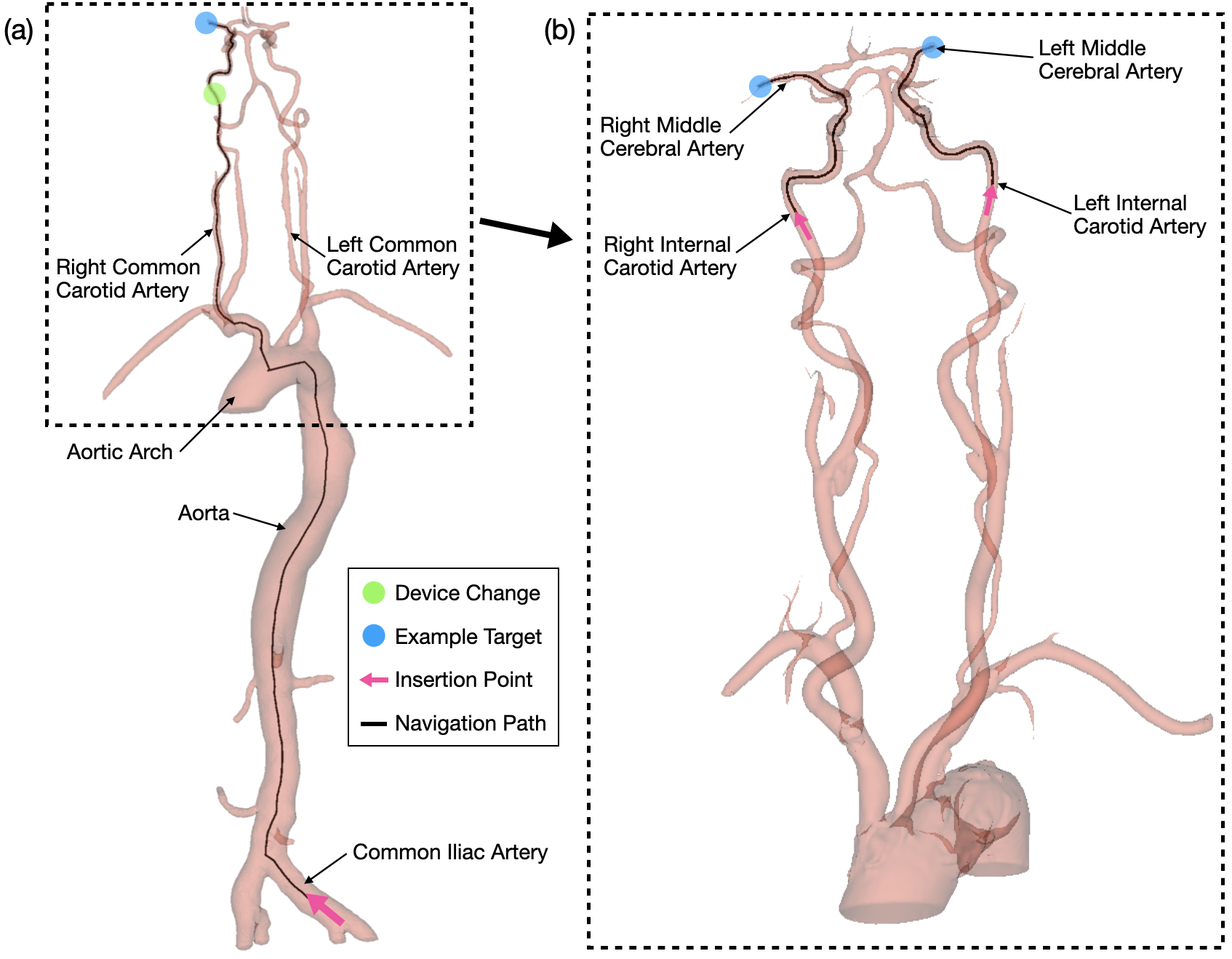}
            % \caption{ MT example of entire navigation path.}
            % \caption{ Micro-guidewire and micro-catheter component of navigation path.}
            \caption{ Example navigation path taken for (a) entire MT navigation path and (b) micro-guidewire and micro-catheter component of navigation path.}
            \label{fig:envs}
        \end{figure}

        The first MT stage of navigating a large guide catheter to form a stable platform has recently been achieved autonomously using RL \cite{Robertshaw2024}. This study simulates the second stage, navigating the micro-guidewire and micro-catheter to the target site (Fig.~\ref{fig:envs}). When these devices are manipulated towards the MCA, navigation challenges include overcoming tortuous anatomy of the ICA and avoiding catheterization of incorrect branches such as the ipsilateral anterior cerebral artery. Rotation of the micro-guidewire (whose tip is curved) is required to avoid catheterization of incorrect branches, while both the tortuous anatomy and rotation cause the devices to interact with one another. 
        
        A target location was randomly sampled from ten centerline points in the right MCA (RMCA) or the left MCA (LMCA), starting from a stable guide catheter platform in the right ICA (RICA) or the left ICA (LICA) cervical segments \cite{Shapiro2014}. Targets in the RMCA could only be navigated from the RICA, and vice-versa for the left side. For each navigation attempt, the target location and the vasculature were changed. 

        The device position was described by three points equally spaced 2\,\unit{\milli\meter} apart along the device. These were denoted as $(x', y')_{i=1,2,3}$, with $(x', y')_{1}$ representing the Euclidean instrument tip co-ordinates. The target location was specified by the target's $(x', y')$-coordinates. Observations comprised current and previous device positions, target location, and the previous action taken.

    \subsection{Simulation environment}

        The \textit{in silico} environment for the navigation task builds on previous work from \cite{Robertshaw2024, Jackson2023, Karstensen2023} and utilized the stEVE framework \cite{Karstensen2024}. The BeamAdapter plugin for Simulation Open Framework Architecture (SOFA) was used to model the Echelon 10 (1.7~F) micro-catheter (Medtronic, Minnesota, USA) and the Synchro 14 micro-guidewire (Stryker NeuroVascular, Cork, Ireland), characterized by Young’s moduli of 47 and 43\,\unit{\mega\pascal}, respectively \cite{Duriez2006, Faure2012}. To represent the rigid distal ICA and the inflexible M1, the simulation assumed rigid vessel walls (with an empty lumen). The simulation's fidelity to real-world guidewire behavior was ensured by utilizing a tensile testing machine to measure the tensile strength of the devices, which facilitated the calculation of its stiffness. While friction between wall and guidewire has been iteratively tuned to mimic guidewire behavior in a test-bench set-up. This is based on previous methodology which has previously allowed \textit{in silico} to \textit{ex vivo} translation of autonomous endovascular navigation designs using models of porcine liver vasculature \cite{Karstensen2022}. 
        
        Input parameters, including guidewire rotation and translation speed, were applied at the proximal device end. The rotational and translational speed were constrained to a maximum of 180\,\unit{\degree\per\second} and 40\,\unit{\mm\per\second}, respectively. Similar to a clinical scenario with fluoroscopy, feedback during the navigation was given as two-dimensional $(x',y')$ tracking coordinates of three points along each device's tip; no visual information showing the geometry of the patient vasculature was given. Experiments were performed using a dual-tracking (combined wire and catheter tracking) method \cite{Robertshaw2024}.

    \subsection{Dataset}

        Twelve computed tomography angiography (CTA) scans which encompassed the aortic arch and extended to the cerebral vessels (obtained with UK Research Ethics Committee 20/ES/0005) were processed into three-dimensional surface meshes using the vascular modeling toolkit and segmentation tools in 3D Slicer \cite{Fedorov2012, Piccinelli2009}. Characteristics of each vasculature can be found in Table~\ref{tab:vascinfo}, and CTA scan parameter settings can be found in the Appendix. These were loaded into SOFA, which allowed augmentation to be applied during training via random scaling (0.7 to 1.3 for height and width) to enhance generalization.

        Demonstrator data was collected \textit{in silico} for two random targets in the RMCA and LMCA across ten patient vasculatures, with four navigation's per vasculature using inputs from a keyboard. The demonstrator was a robotics engineer with 2-years' experience using \textit{in silico} navigation environments and \textit{in vitro} endovascular robots, with 2 years of tuition from an expert interventional neuroradiologist (UK consultant; US attending equivalent). Data collected included actions (rotation and translation), device tracking, and target location. Data was split by left or right target navigation.

    \subsection{Controller architecture}

\begin{figure}[]
            \centering
            \includegraphics[width=9cm]{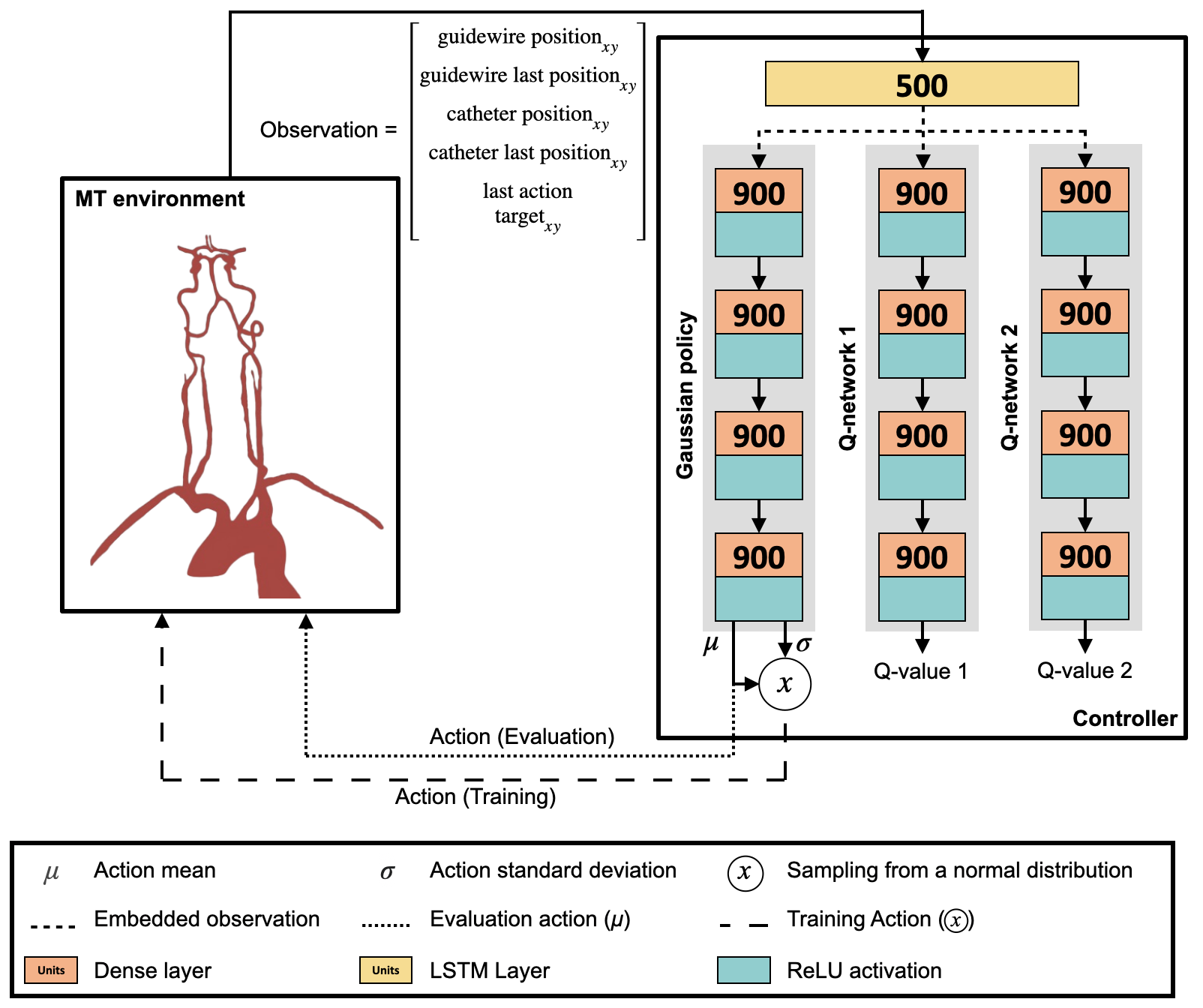}
            \caption{ Architecture of the proposed controller based on a SAC design. Reward and action input for the Q-networks are not included in this diagram. Adapted from \cite{Karstensen2023}.}
            \label{fig:SACalgo}
        \end{figure}

        All RL models in this study were trained using a Soft Actor-Critic (SAC) controller, implemented using PyTorch \cite{Paszke2019}. This is shown in Fig.~\ref{fig:SACalgo}, and is adapted from previous work \cite{Robertshaw2024, Karstensen2023}. The architecture includes a Long Short-Term Memory layer for learning trajectory-dependent state representations, which has been shown to allow probing of the correct vessel when the target branch is not unambiguously located from the target coordinates \cite{Karstensen2023}. The architecture also includes feedforward layers for controlling the guidewire. The controller takes observations as input, and a Gaussian policy network outputs mean ($\mu$) and standard deviation ($\sigma$) for expected actions, representing the micro-catheter's and micro-guidewire's rotation and translation. During training, actions are sampled from the $\sigma$, but for evaluation, $\mu$ is used directly for deterministic behavior. 

        Maximum entropy IRL was employed to learn reward functions from demonstrations for navigating vascular branches, aiming to replicate expert behavior while maximizing entropy to account for variability \cite{Ziebart2008}. This method, suited for complex environments like endovascular navigation, can outperform methods such as behavioral cloning and standard RL, which may overfit or struggle with sparse rewards and has provided improvements over the state-of-the-art in autonomous endovascular interventions \cite{Robertshaw2024}. A feedforward neural network with four fully connected layers was used in the IRL model, trained with expert trajectories. Training of the IRL model involved 1~million iterations with a learning rate of $1 \times 10^{-2}$, entropy regularization ($\alpha = 0.1$), and a 0.01 length penalty. Separate IRL models were trained for each vascular branch, with the ipsilateral model returning reward values for the current observations.

    \subsection{Reward functions}
    
        Six reward functions were used to assess whether adding micro-guidewire tip force feedback improved \textit{in silico} results. Each reward function was calculated in every simulation step based on the agent's actions. Therefore, actions led to different reward quantities across reward functions and, hence, a variation in the final model. The first three reward functions followed prior work \cite{Robertshaw2024}:
        
        \begin{itemize}
            \item $R_1$ (Eq.~\ref{eq:R1}): Dense reward, where \textit{pathlength} is the distance between the guidewire tip and the target, and $\Delta\text{pathlength}$ is the change from the previous step.
            \begin{equation}
                R_{1} = -0.005 - 0.001\cdot\Delta\text{pathlength}+\begin{cases}1.0 & \text{if target reached}\\0 & \text{else}\end{cases}
                \label{eq:R1}
            \end{equation}
            \item $R_2$ (Eq.~\ref{eq:R2}): IRL-derived reward, where $R_{RICA}$ or $R_{LICA}$ is calculated based on the carotid artery the target is in.
            \begin{equation}
                R_{2} = \begin{cases}
                    R_{RICA} & \text{if target in RICA}\\
                    R_{LICA} & \text{if target in LICA}
                \end{cases}
                \label{eq:R2}
            \end{equation}
            \item $R_3$ (Eq.~\ref{eq:R3}): A combined reward model of $R_1$ and $R_2$, scaled by $\alpha = 0.001$.
            \begin{equation}
                R_{3} = R_{1} +  \alpha R_{2}
                \label{eq:R3}
            \end{equation}
        \end{itemize}

        Three additional reward functions ($R_4$, $R_5$, and $R_6$) were created by adding force feedback to $R_1$, $R_2$, and $R_3$, achieved by implementing a collision monitor on the micro-guidewire tip, to give $P_{x,y,z}$ at each simulation step. Analysis of demonstrator data guidewire tip forces showed that $\|P\|$ values remained at 0.80\,\unit{\newton} or less, therefore if $\|P\|$ was greater than a threshold value of 0.85\,\unit{\newton}, a negative reward was given. This was added to $R_{1}$, $R_{2}$ and $R_{3}$, to give $R_{4}$, $R_{5}$ and $R_{6}$, as shown in Equation~\ref{eq:Rj}. The reward functions were fine-tuned iteratively through RL training by systematically adjusting the reward weights and $\alpha$, with each adjustment followed by evaluation on key metrics of success rate, procedure time, and tip force reduction. As guidewire tip forces are only used in the reward function during training, the agent does not require a force input during \textit{in silico} evaluation, or any experimental stages beyond this (e.g., \textit{in vitro} evaluation).

        \begin{equation}
            R_{j} = R_{j-3}-0.01\cdot\begin{cases}P-0.85 & \text{if }\|P\|>0.85 \\0 & \text{if }\|P\|\leq0.85\end{cases} \quad \text{for } j = 4, 5, 6
            \label{eq:Rj}
        \end{equation}

    \subsection{Implementation}
        
         Training consisted of $1 \times 10^7$ exploration steps, with each navigation task (or episode) considered complete when the target was reached within a 5~mm threshold, which is suitable to cover the entire cross-section of the vessel. A timeout of 200 steps (approximately 27\,\unit{\second}) was set for computational efficiency. Training was performed on an NVIDIA DGX A100 node (Santa Clara, California, USA) with 8 GPUs, and took approximately 48 hours per model.

    \subsection{Experimental design and evaluation}\label{sec:eval}

\begin{figure}[]
            \centering
            \includegraphics[width=9cm]{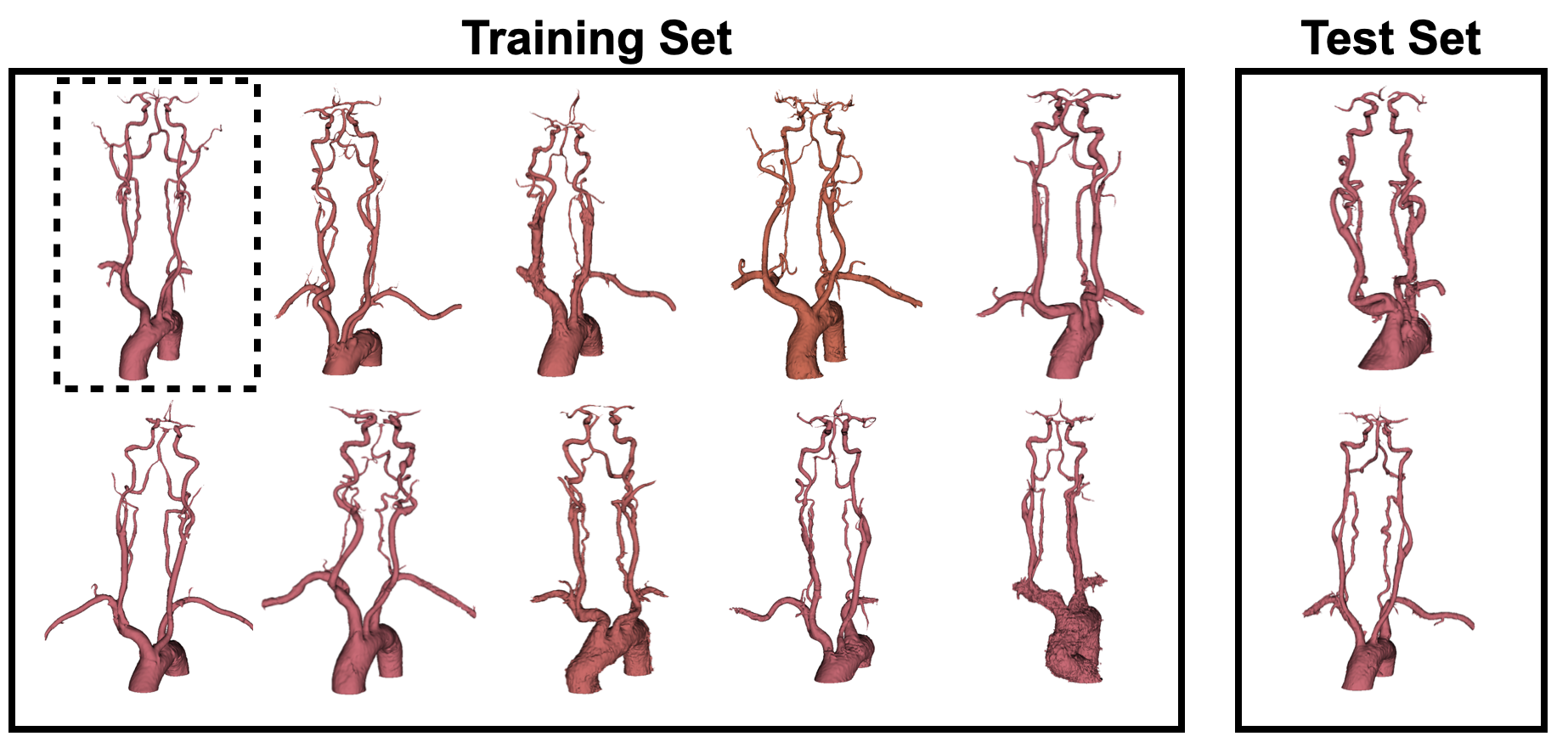}
            \caption{ Patient vasculatures used for training (10 cases) and testing (2 cases). The dashed box represents the single vasculature used for training and testing in the first experiment (Section \ref{sec:eval}).}
            \label{fig:meshes}
        \end{figure}

        Two experiments were conducted. First, a single patient vasculature was used for both training and testing to replicate and confirm that a combined reward model provides improved success rates over a traditional dense reward function, as in \cite{Robertshaw2024}, for the second stage of MT. Second, ten different patient vasculatures were used during training with augmentation applied, while two were reserved for testing only (Fig.~\ref{fig:meshes}). This created a train/test split of approximately 80/20 \cite{Gholamy2018}.

        For the first experiment, five targets per branch were used for training and five different targets for evaluation. For the second experiment, all ten targets in each branch were utilized for evaluation. Evaluations were conducted every $2.5 \times 10^5$~exploration steps for 80~episodes, recording the success rate, procedure time, and path ratio \cite{Robertshaw2024}. The models with the highest success rate for each reward function were compared against one another for path ratio, procedure time, mean forces, and exploration steps. Comparative statistical analyses were conducted using two-tailed paired Student's t-tests and Analysis of Variance, with a predetermined significance threshold set at $p = 0.05$.

\section{Results}

    \subsection{Training and testing on the same patient vasculature}

        Table~\ref{tab:mesh4table} presents the results of training and testing with reward functions $R_{1-6}$ on the same patient vasculature. Both $R_{1}$ (dense without force-feedback) and $R_{3}$ (combined model without force-feedback) reach success rates of 100\%, while $R_{4}$ (dense with force-feedback) and $R_{6}$ (combined model with force-feedback) reach 97\% and 96\%, respectively ($R_{1,4}$: $p= 0.157$, $R_{3,6}$: $p= 0.081$). Both $R_{2}$ (IRL without force-feedback) and $R_{5}$ (IRL with force-feedback) had a 44\% success rate. The addition of force feedback was associated with significantly longer procedure times for $R_{2}$ to $R_{5}$ (8.1\,s to 15.7\,s [$p= 0.001$]) and $R_{3}$ to $R_{6}$ (5.0\,s to 6.9\,s [$p< 0.001$]), and significantly lower mean forces for $R_{1}$ to $R_{4}$ (0.29~\si{\newton} to 0.25~\si{\newton} [$p< 0.001$]) and $R_{2}$ to $R_{5}$ (0.30~\si{\newton} to 0.28~\si{\newton} [$p= 0.044$]). Procedural time for $R_{1}$ to $R_{4}$ (3.8\,s to 3.8\,s [$p= 0.794$]) and mean force for $R_{3}$ to $R_{6}$ (0.25~\si{\newton} to 0.24~\si{\newton} [$p= 0.249$]) were both not significantly different. The maximum force reported across all evaluations was 1.0~\si{\newton} or less.

        \begin{table*}[]
            \footnotesize
            \centering
            \caption{Results of testing and training on the same patient vasculature. Values are reported as mean $\pm$ standard deviation. \textit{Success Rate}: percentage of evaluation episodes where target is reached. \textit{Path Ratio}: percentage distance navigated to target in unsuccessful episodes, calculated by dividing the total distance navigated toward the target by the initial distance. \textit{Procedure Time}: time from the start of navigation to the target location for successful episodes. \textit{Mean Force}: mean magnitude of tip forces during evaluation. \textit{Exploration Steps}: number of training steps taken to reach the point at which the results are provided. The reported values are mean $\pm$ standard deviation (standard deviation values may exceed logical bounds (0–100\%) due to the statistical calculation).}
            \label{tab:mesh4table}
            \begin{tabular}{l|l|l|l|l|l}
                \multicolumn{1}{c|}{\textbf{\begin{tabular}[c]{@{}c@{}}Reward\\ Function\end{tabular}}} &
                  \multicolumn{1}{c|}{\textbf{\begin{tabular}[c]{@{}c@{}}Success\\ Rate (\%)\end{tabular}}} &
                  \multicolumn{1}{c|}{\textbf{\begin{tabular}[c]{@{}c@{}}Procedure\\ Time (s)\end{tabular}}} &
                  \multicolumn{1}{c|}{\textbf{\begin{tabular}[c]{@{}c@{}}Path \\ Ratio (\%)\end{tabular}}} &
                  \multicolumn{1}{c|}{\textbf{\begin{tabular}[c]{@{}c@{}}Mean \\ Force (N)\end{tabular}}} &
                  \multicolumn{1}{c}{\textbf{\begin{tabular}[c]{@{}c@{}}Exploration \\ Steps\end{tabular}}} \\ \hline
                \multicolumn{1}{c|}{$R_{1}$} & \multicolumn{1}{c|}{$100 \pm 0$} & \multicolumn{1}{c|}{$3.8 \pm 0.3$} & \multicolumn{1}{c|}{$100 \pm 0$} & \multicolumn{1}{c|}{$0.29 \pm 0.37$} & \multicolumn{1}{c}{$325 \times 10^6$}\\ \hline
                \multicolumn{1}{c|}{$R_{2}$} & \multicolumn{1}{c|}{$44 \pm 50$} & \multicolumn{1}{c|}{$8.1 \pm 8.6$} & \multicolumn{1}{c|}{$75.1 \pm 23.8$} & \multicolumn{1}{c|}{$0.30 \pm 0.37$} & \multicolumn{1}{c}{$125 \times 10^6$}\\ \hline
                \multicolumn{1}{c|}{$R_{3}$} & \multicolumn{1}{c|}{$100 \pm 0$} & \multicolumn{1}{c|}{$5.0 \pm 2.1$} & \multicolumn{1}{c|}{$100 \pm 0$} & \multicolumn{1}{c|}{$0.25 \pm 0.36$} & \multicolumn{1}{c}{$300 \times 10^6$}\\ \hline
                \multicolumn{1}{c|}{$R_{4}$} & \multicolumn{1}{c|}{$97 \pm 16$} & \multicolumn{1}{c|}{$3.8 \pm 1.0$} & \multicolumn{1}{c|}{$96 \pm 4.5$} & \multicolumn{1}{c|}{$0.25 \pm 0.36$} & \multicolumn{1}{c}{$175 \times 10^6$}\\ \hline
                \multicolumn{1}{c|}{$R_{5}$} & \multicolumn{1}{c|}{$44 \pm 50$} & \multicolumn{1}{c|}{$15.6 \pm 9.4$} & \multicolumn{1}{c|}{$80.0 \pm 18.9$} & \multicolumn{1}{c|}{$0.28 \pm 0.36$} & \multicolumn{1}{c}{$75 \times 10^6$}\\ \hline
                \multicolumn{1}{c|}{$R_{6}$} & \multicolumn{1}{c|}{$96 \pm 19$} & \multicolumn{1}{c|}{$6.9 \pm 2.4$} & \multicolumn{1}{c|}{$94.6 \pm 7.3$} & \multicolumn{1}{c|}{$0.24 \pm 0.35$} & \multicolumn{1}{c}{$350 \times 10^6$}\\
            \end{tabular}
        \end{table*}

    \subsection{Testing on unseen patient vasculatures}

\begin{figure}[]
            \includegraphics[width=13cm]{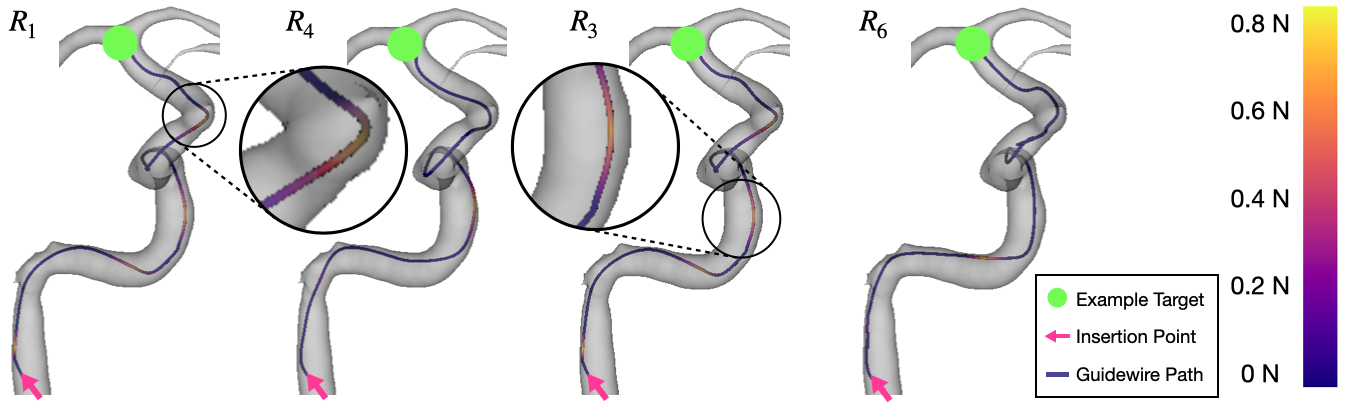}
            \centering
            \caption{ Navigation path to RMCA for $R_{1}$ (dense without force-feedback), $R_{3}$ (combined model without force-feedback), $R_{4}$ (dense with force-feedback), and $R_{6}$ (combined model with force-feedback), with example regions of higher tip forces indicated.}
            \label{fig:nav}
        \end{figure}

        Table~\ref{tab:multitable} shows the results of testing reward functions $R_{1}$, $R_{3}$, $R_{4}$, and $R_{6}$ on two unseen patient vasculatures. Overall, there was slight degradation in outcomes except for mean force when compared to the initial `single vasculature training/testing' outcomes. A notable exception was $R_6$, which showed that when testing on unseen vasculatures was compared to the initial `single vasculature training/testing' scenario, results remained consistent across success rate (96\% to 96\% [$p= 0.987$]), procedure time (6.9\,s to 7.0\,s [$p= 0.754$]), path ratio (94.6\% to 93.5\% [$p= 0.463$]), and mean force (0.24~\si{\newton} to 0.24~\si{\newton} [$p= 0.947$]). $R_6$ achieved the highest success rate of 96\%, followed by 93\% for $R_1$ ($p=0.306$). In contrast to the initial `single vasculature training/testing' scenario, force-feedback led to decreased procedure times for both $R_1$ to $R_4$ (8.1\,s to 4.5\,s, $p=0.004$), and $R_3$ to $R_6$  (13.6\,s to 7.0\,s, $p< 0.001$). Mean forces decreased significantly when force feedback was included for $R_1$ to $R_4$ (0.29~\si{\newton} to 0.24~\si{\newton} [$p< 0.001$]), and $R_3$ to $R_6$ (0.27~\si{\newton} to 0.24~\si{\newton} [$p= 0.017$]). Illustrative differences in tip forces during navigation for $R_1$ and $R_6$ can be seen in Fig.~\ref{fig:nav}. The maximum force reported across all evaluations was 1.0~\si{\newton} or less.

        \begin{table*}[]
            \footnotesize
            \centering
            \caption{Results of testing on two unseen vasculatures after training on ten different vasculatures. Table~\ref{tab:mesh4table} concluded IRL’s ineffectiveness compared to other reward functions, and hence $R_{2, 5}$ were omitted when testing on unseen vasculatures.}
            \label{tab:multitable}
            \begin{tabular}{l|l|l|l|l|l}
                \multicolumn{1}{c|}{\textbf{\begin{tabular}[c]{@{}c@{}}Reward\\ Function\end{tabular}}} &
                  \multicolumn{1}{c|}{\textbf{\begin{tabular}[c]{@{}c@{}}Success\\ Rate (\%)\end{tabular}}} &
                  \multicolumn{1}{c|}{\textbf{\begin{tabular}[c]{@{}c@{}}Procedure\\ Time (s)\end{tabular}}} &
                  \multicolumn{1}{c|}{\textbf{\begin{tabular}[c]{@{}c@{}}Path \\ Ratio (\%)\end{tabular}}} &
                  \multicolumn{1}{c|}{\textbf{\begin{tabular}[c]{@{}c@{}}Mean \\ Force (N)\end{tabular}}} &
                  \multicolumn{1}{c}{\textbf{\begin{tabular}[c]{@{}c@{}}Exploration \\ Steps\end{tabular}}} \\ \hline
                 \multicolumn{1}{c|}{$R_{1}$}& \multicolumn{1}{c|}{$93 \pm 27$} & \multicolumn{1}{c|}{$8.1 \pm 8.8$} & \multicolumn{1}{c|}{$89.6 \pm 23.8$} & \multicolumn{1}{c|}{$0.29 \pm 0.37$} & \multicolumn{1}{c}{$400 \times 10^6$} \\ \hline
                 \multicolumn{1}{c|}{$R_{3}$} & \multicolumn{1}{c|}{$88 \pm 33$} & \multicolumn{1}{c|}{$13.6 \pm 10.4$} & \multicolumn{1}{c|}{$88.1 \pm 23.6$} & \multicolumn{1}{c|}{$0.27 \pm 0.36$} & \multicolumn{1}{c}{$625 \times 10^6$} \\ \hline
                 \multicolumn{1}{c|}{$R_{4}$} & \multicolumn{1}{c|}{$78 \pm 42$} & \multicolumn{1}{c|}{$4.5 \pm 3.0$} & \multicolumn{1}{c|}{$84.1 \pm 28.7$} & \multicolumn{1}{c|}{$0.24 \pm 0.35$} & \multicolumn{1}{c}{$200 \times 10^6$} \\ \hline
                 \multicolumn{1}{c|}{$R_{6}$} & \multicolumn{1}{c|}{$96 \pm 19$} & \multicolumn{1}{c|}{$7.0 \pm 3.9$} & \multicolumn{1}{c|}{$93.5 \pm 15.1$} & \multicolumn{1}{c|}{$0.24 \pm 0.35$} & \multicolumn{1}{c}{$700 \times 10^6$}
            \end{tabular}
        \end{table*}

\section{Discussion}

    This study is the first to achieve autonomous navigation of both micro-guidewire and micro-catheter during the second stage of MT, from the ICA to the MCA. It builds upon previous research by evaluating the reward function used by \cite{Robertshaw2024}, while testing on unseen patient vasculatures. This proof-of-concept \textit{in silico} study also demonstrated that incorporating tip force feedback into the reward function is feasible in the MT second stage scenario shown here, and may reduce mean forces on vasculature. Furthermore, when force feedback is incorporated into a combined reward model and tested on unseen vasculatures, improved success rates and reduced procedure times are seen. The same model appeared generalizable with no domain shift when transitioning from training/testing on a single vasculature, to training on multiple vasculatures and testing on unseen ones. While the current autonomous MT navigation system is at technology readiness level (TRL) 3 \cite{Mankins1995}, the first successful autonomous navigation of the second stage of MT on unseen real patient anatomies might contribute towards increasing the TRL; a step forward to realizing the benefits of fully autonomous MT.

    \subsection{IRL reward function}

        Results from training and testing on the same vasculature aligned with previous work \cite{Robertshaw2024}, and showed high success rates for $R_{1,4}$ and $R_{3,6}$, while $R_2$ had a lower success rate. These initial results concluded IRL's ineffectiveness compared to the other reward functions, leading to the omission of $R_2$ and $R_5$ in subsequent testing on unseen vasculatures. This ineffectiveness may stem from the inherent difficulty in capturing the nuances of expert decision-making in highly dynamic environments like endovascular navigation, where factors such as subtle anatomical differences and force-feedback cues play a role. 
        
        Training RL models, particularly in IRL settings, remains a considerable challenge. The dual-task of learning a reward function from demonstrations and optimizing a policy introduces additional complexity. Our findings highlight the sensitivity of RL approaches to the reward function, which impact convergence speed as exploration steps increased with the addition of IRL when moving from $R_1$ to $R_3$ ($400\times10^6$ to $625\times10^6$), and from $R_4$ to $R_6$ ($200\times10^6$ to $700\times10^6$).
        
        Despite this, expert demonstrator data could plausibly still be useful in enhancing autonomous endovascular navigation, whether this be through a different form of IRL or alternative RL algorithms. Nevertheless, the addition of demonstrator data has been shown to improve RL in different scenarios, while enhancing sample efficiency, helping with sparse rewards, and potentially enabling safer exploration \cite{Ramirez2022}, and has effectively been utilized in previous autonomous endovascular navigation work \cite{Kweon2021}.

    \subsection{Force feedback}

        A decrease in success rate, and path ratio, as well as an increase in procedure time, and training time, was observed when moving from testing and training on the same vasculature to testing on unseen vasculatures. While this overall slight degradation in outcomes is expected due to the difficult task of navigating a completely unseen vasculature there were two exceptions. First, mean forces were lower demonstrating safety. Second, a maximum success rate of 96\% is promising for the combined model. Whilst the addition of guidewire tip force feedback in the reward function decreased success rate for the dense reward function (93\% to 78\%), the addition improved the success rate for the combined model (88\% to 96\%). As there are multiple reward signals for all reward functions examined (tip force, steps taken, path length), the reason for this difference may be that $R_6$ provides the optimal balance of these signals. By fine-tuning the scaling factors for each reward function, it may be possible to further improve training results. Force feedback also significantly reduced procedure time for unseen vasculatures ($R_{1,4}$: 8.1s to 4.5s, $R_{3,6}$: 13.6s to 7.0s) - a result not seen when testing and training on the same vasculature. Plausibly, the addition of forces may discourage the tip from applying significant force to the vessel wall which can occur when the tip is caught within the lumen preventing it from moving; the net result might be an increase in overall velocity due to decreased friction, and subsequently decrease procedure time.

        Our motivation for including force feedback was to decrease mean forces as a patient safety feature. While its addition decreased the mean forces on the guidewire tip, we acknowledge that the mean forces across all experiments in unseen vasculature were low in any case with a maximum of 0.29~\si{\newton}, well below the vessel rupture threshold of 1.5~\si{\newton} proposed by \cite{Jackson2023} Although no study has analyzed instrument forces during RL training as a method for increasing safety during the procedure, previous studies have measured instrument and vessel wall forces as part of their experiments \cite{Chi2020}, which suggests that our approach could be translated to \textit{in vitro} environments.
        
    \subsection{Limitations}

        Similar to previous methods \cite{Robertshaw2024, Jackson2023, Karstensen2023}, \textit{in silico} work cannot measure fluoroscopy time (a surrogate of radiation exposure) which has been included in some clinical studies \cite{Beaman2023}. Additionally, using a keypad-operated controller for data collection may not fully reflect the actions of an experienced neurointerventional radiologist. The small dataset of patient vasculatures used for training and testing limits generalizability. Although augmentation introduced a degree of variability, this dataset is not able to replicate the different configurations of the Circle of Willis or branching patterns of the cerebral arteries found in the population, and a larger, more diverse, dataset may be required to ensure effective \textit{in vitro} translation. As the current work focuses entirely \textit{in silico}, physical characteristics of guidewires, especially force transmission, may not be fully captured. Hence, future work should aim to understand the impact that training with \textit{in silico} force feedback has on  \textit{in vitro} models, while providing \textit{in vitro} validation of the proposed methods. While this work incorporated real-world mechanical data to enhance the accuracy of force models, future work could improve simulation-to-real transfer by modeling vessels as elastic bodies to reflect the behavior of real vessels - however, any added benefit may be imperceptible during the second MT stage used in the current scenario, as the distal ICA is rigid and the M1 somewhat inflexible. Future work may also compare tracking- and visual-based RL algorithms, to investigate whether the same benefits can be gained from providing the entire image frame at each simulation step.       
    
\section{Conclusion}

    In conclusion, this study proposes an offline RL algorithm that autonomously navigates (micro-guidewires and micro-catheters) to cerebral vessels in unseen patients while considering the amount of force applied to the device tip, representing the first demonstration of autonomous navigation in both cerebral vessels and unseen patient vasculatures. We evaluated various reward functions and confirmed the superior performance of a combined reward model with force-feedback ($R_6$), which maintained a 96\% success rate in patient vasculatures the model had not seen previously. The addition of force-feedback not only reduced procedure times but also decreased the mean forces exerted, plausibly enhancing patient safety by decreasing both time to treatment as well as the likelihood of iatrogenic vessel damage. Future work should focus on translating the proposed method to \textit{in vitro} experiments and extending the dataset to integrate more diverse patient vasculatures.

\section*{Acknowledgments}
The authors would like to thank Han-Ru Wu (Nuffield Department of Primary Care Health Sciences, University of Oxford, UK) for her work on computing the characteristics for the vasculatures used. 

\section*{Funding}
Partial financial support was received from the WELLCOME TRUST (Grant Agreement No 203148/A/16/Z) and the Engineering and Physical Sciences Research Council Doctoral Training Partnership (Grant Agreement No EP/R513064/1). For the purpose of Open Access, the Author has applied a CC BY public copyright license to any Author Accepted Manuscript version arising from this submission.

\section*{Appendix}
    \subsection*{Vasculature characteristics}
        \begin{table*}[htb!]
            \footnotesize
            \centering
            \caption{Characteristics of vasculatures used during the study \cite{Lahlouh2023}. Tortuosity is computed as the ratio between two distances: (1) the geodesic length along the centerline of the artery, and (2) the Euclidean distance between the two end-points of the artery segment. This index lies in $[1, +\infty)$: the closer to 1 the less tortuous. Radius was calculated as the origin.}
            \label{tab:vascinfo}
            \begin{tabular}{c|c|c|c|c|c|c|c}
            \textbf{\begin{tabular}[c]{@{}c@{}}Vasculature\\ No.\end{tabular}} & \textbf{\begin{tabular}[c]{@{}c@{}}Train/\\ test\end{tabular}} & \textbf{\begin{tabular}[c]{@{}c@{}}RICA\\ tortuosity\end{tabular}} & \textbf{\begin{tabular}[c]{@{}c@{}}LICA\\ tortuosity\end{tabular}} & \textbf{\begin{tabular}[c]{@{}c@{}}RICA\\ radius\\ (mm)\end{tabular}} & \textbf{\begin{tabular}[c]{@{}c@{}}RMCA\\ radius\\ (mm)\end{tabular}} & \textbf{\begin{tabular}[c]{@{}c@{}}LICA\\ radius\\ (mm)\end{tabular}} & \textbf{\begin{tabular}[c]{@{}c@{}}LMCA\\ radius\\ (mm)\end{tabular}} \\ \hline
            1 & Train & 1.16 & 1.14 & 2.05 & 1.21 & 1.51 & 2.18 \\ \hline
            2 & Train & 1.25 & 1.18 & 2.13 & 1.28 & 2.48 & 1.15 \\ \hline
            3 & Train & 1.35 & 1.42 & 3.46 & 1.00 & 3.30 & 0.77 \\ \hline
            4 & Train & 1.16 & 1.13 & 0.99 & 1.19 & 3.66 & 1.19 \\ \hline
            5 & Train & 1.10 & 1.24 & 3.80 & 1.66 & 2.49 & 1.58 \\ \hline
            6 & Train & 1.19 & 1.27 & 3.10 & 1.40 & 1.33 & 1.02 \\ \hline
            7 & Train & 1.47 & 1.60 & 2.87 & 1.78 & 3.56 & 1.59 \\ \hline
            8 & Train & 1.41 & 1.33 & 3.27 & 1.13 & 2.83 & 1.10 \\ \hline
            9 & Train & 1.39 & 1.23 & 2.36 & 1.61 & 4.26 & 1.51 \\ \hline
            10 & Train & 1.27 & 1.42 & 3.51 & 1.64 & 3.52 & 1.44 \\ \hline
            11 & Test & 1.15 & 1.11 & 5.80 & 0.98 & 5.35 & 0.93 \\ \hline
            12 & Test & 1.21 & 1.21 & 3.33 & 0.98 & 3.20 & 0.84
            \end{tabular}
        \end{table*}

    \subsection*{CTA Scan Parameter Settings}\label{CTscans}
        \subsubsection*{Scan Protocol}
            \begin{itemize}
                \item \textbf{Scanner Type:} Multi-detector CT (Optima 660 GE Healthcare, 64-section)
                % \item \textbf{Scan Range:} From the aortic arch to the circle of Willis (covering the extracranial and intracranial arteries)
            \end{itemize}
        \subsubsection*{Acquisition Parameters}
            \begin{itemize}
                \item \textbf{Slice Thickness:} 0.625~\unit{\mm}
                \item \textbf{Detector Collimation:} 40.0~\unit{\mm}
                \item \textbf{Pitch:} 0.984~\unit{\mm}
                \item \textbf{Rotation Time:} 0.3-0.5\,\unit{\second} (faster for reducing motion artifacts)
                \item \textbf{Field of View (FOV):} 500~\unit{\mm}
            \end{itemize}
        \subsubsection*{Contrast Injection}
            \begin{itemize}
                \item The acute-stroke imaging protocol consisted of craniocervical arterial phase acquisition after intravenous injection of 50~\unit{\milli\liter} of iohexol, 647~\unit{\milli\gram\per\milli\liter} (5~\unit{\milli\liter\per\second}) (Omnipaque 300; GE Healthcare)
                % \item \textbf{Contrast Type:} Non-ionic iodinated contrast (e.g., Iohexol, Iopamidol)
                % \item \textbf{Volume:} 50 to 80~\unit{\milli\liter} (depending on patient size)
                % \item \textbf{Injection Rate:} 4 to 6~\unit{\milli\liter\per\second} (high flow rate for arterial phase enhancement)
                \item \textbf{Saline Flush:} 40~\unit{\milli\liter} (saline bolus)
                \item \textbf{Bolus Tracking:} Self triggered - observe contrast and scan when suitable contrast in ascending aorta and carotid artery plus auto minimum delay setting (15-20~\unit{\second} post-injection for optimal arterial phase)
            \end{itemize}
        % \subsection{Imaging Phases}
        %     \begin{itemize}
                % \item \textbf{Timing:} Arterial phase imaging is crucial. Scans typically start 5-10~\unit{\second} after contrast bolus tracking triggers
                % \item \textbf{Scan Delay (if manual):} 15-20~\unit{\second} post-injection for optimal arterial phase
            % \end{itemize}
        \subsubsection*{Radiation Dose Parameters}
            \begin{itemize}
                \item \textbf{Tube Voltage (kVp):} 100~kVp
                \item \textbf{Tube Current (mA):} 315~\unit{\milli\ampere} with dose reduction on 40\%
                % \item \textbf{Dose Modulation:} Activated to optimize radiation dose and maintain image quality
                \item \textbf{DLP (Dose Length Product):} Towards 600~\unit{\milli\gray\cm}
            \end{itemize}
        \subsubsection*{Reconstruction Parameters}
            \begin{itemize}
                \item \textbf{Reconstruction Algorithm:} Standard kernel for arteries 
                % \item \textbf{Reconstruction Interval:} 0.3 to 0.5~\unit{\mm} overlap for high-resolution 3D reconstructions
                \item \textbf{Post-processing:} 2/5 Maximum intensity projection (MIP) 
            \end{itemize}
        \subsubsection*{Patient Positioning}
            \begin{itemize}
                \item Supine position, head immobilized, and neck slightly extended to avoid venous overlap with arteries
            \end{itemize}

%Bibliography
\bibliographystyle{unsrt}  
\bibliography{references}

\end{document}